\documentclass{ecai}
\usepackage{times}  
\usepackage{array}
\usepackage{mathtools}
\usepackage{multirow}
\usepackage{enumitem}
\usepackage{dsfont}
\usepackage{url}  
\usepackage{graphicx, subfigure}
\usepackage{amssymb}  
\usepackage[linesnumbered,boxed,ruled]{algorithm2e}

\newtheorem{Definition}{Definition}
\newtheorem{Theorem}{Theorem}

\newcommand{\nosection}[1]{\vspace{2pt}\noindent\textbf{#1.}}
\begin{document}
\title{Secure Social Recommendation based on Secret Sharing}
\author{
Chaochao Chen\institute{Ant Financial Services Group, China, email: chaochao.ccc@antfin.com}~, 
Liang Li\institute{Huawei Noah's Ark Lab, China, email: liliang103@huawei.com}~, 
Bingzhe Wu\institute{Peking University, China, email: wubingzhe@pku.edu.cn}~, 
Cheng Hong\institute{Alibaba security, China, email: vince.hc@alibaba-inc.com}~,  
Li Wang\institute{Ant Financial Services Group, China, email: raymond.wangl@antfin.com}~, 
Jun Zhou\institute{Ant Financial Services Group, China, email: jun.zhoujun@antfin.com}~
}
\maketitle

\begin{abstract}
Nowadays, privacy preserving machine learning has been drawing much attention in both industry and academy. Meanwhile, 
recommender systems have been extensively adopted by many commercial platforms (e.g. Amazon) and they are mainly built based on user-item interactions. 
Besides, social platforms (e.g. Facebook) have rich resources of user social information. 
It is well known that social information, which is rich on social platforms such as Facebook, are useful to build intelligent recommender systems. 
It is anticipated to combine the social information with the user-item ratings to improve the overall  recommendation performance. 
Most existing recommendation models are built based on the assumptions that the social information are available. 
However, different platforms are usually reluctant to (or can not) share their data due to certain concerns. 
In this paper, we first propose a \texttt{SE}cure \texttt{SO}cial \texttt{REC}ommendation (\texttt{SeSoRec}) framework which is able to (1) collaboratively mine knowledge from social platform to improve the recommendation performance of the rating platform, and (2) securely keep the raw data of both platforms. 
We then propose a \texttt{S}ecret \texttt{S}haring based \texttt{M}atrix \texttt{M}ultiplication (\texttt{SSMM}) protocol to optimize \texttt{SeSoRec} and prove its correctness and security theoretically. 
By applying minibatch gradient descent, \texttt{SeSoRec} has linear time  complexities in terms of both computation and communication. 
The  comprehensive experimental results  on three real-world datasets demonstrate the effectiveness of our proposed \texttt{SeSoRec} and \texttt{SSMM}. \end{abstract}

\section{Introduction}

Nowadays, recommender systems have been extensively used in many commercial platforms \cite{chen2018distributed}. 
The key point for recommendation is to use as much information as possible to learn better preferences of users and items. 
To achieve this, besides user-item interaction information, additional information such as social relationship and contextual information have been utilized \cite{ma2011recommender,rendle2010factorization,chen2018semi}. 

Existing researchers usually make the assumption that all kinds of information are available, which is somehow inconsistent with most of the real-world cases.
In practice, different kinds of information are located on different platforms, e.g., huge user-item interaction information on Amazon while rich user social information on Facebook. 
However, different platforms are reluctant to (or can not) share their own data due to competition or regulation reasons. 

Therefore, for the recommendation platforms who have rich user-item interaction data, how to use the additional data such as social information on other platforms to further improve recommendation performance, meanwhile protect the raw data security of both platforms, is a crucial question to be answered. 
It is worthwhile to study such a research topic in both industry and academia. 

Secure Multi-Party Computation (MPC) provides a solution to the above question. 
MPC aims to jointly compute a function for multi-parties while keeping the individual inputs private \cite{yao1986generate}, and 
it has been adopted by many machine learning algorithms for secure data mining, including decision tree \cite{lindell2005secure}, linear regression \cite{nikolaenko2013privacy}, and logistic regression \cite{mohassel2017secureml}. 
However, it has not been applied to the above-mentioned secure multi-party recommendation problems yet. 

 
In this paper, we consider the scenarios where user-item interaction information and user social information are on different platforms, which is quite common in practice. 
Platform $\mathcal{A}$ has user-item interaction information and Platform $\mathcal{B}$ has user social information, the challenge is to improve the recommendation performance of $\mathcal{A}$ by securely using the user social information on $\mathcal{B}$.
To fulfill this, we formalize secure social recommendation as a MPC problem and propose a \texttt{SE}cure \texttt{SO}cial \texttt{REC}ommendation (\texttt{SeSoRec}) framework for it. 
Our proposed \texttt{SeSoRec} is able to 
(1) collaboratively mine knowledge from social platform to improve the recommendation performance of the rating platform, and 
(2) keep the raw data of both platforms securely. 
We further propose a novel \texttt{S}ecret \texttt{S}haring based \texttt{M}atrix \texttt{M}ultiplication (\texttt{SSMM}) protocol to optimize \texttt{SeSoRec}, and we also prove its correctness and security. 
Our proposed \texttt{SeSoRec} and \texttt{SSMM} have linear computation and communication complexities. 
Experimental results conducted on three real-world datasets demonstrate the effectiveness of our proposed \texttt{SeSoRec} and \texttt{SSMM}. 

We summarize our main contributions as follows: 
\begin{itemize}[leftmargin=*] \setlength{\itemsep}{-\itemsep}
\item We observe a secure social recommendation problem in practice, formalize it as a MPC problem, and propose a \texttt{SeSoRec} framework for it. 
\item We propose a novel \texttt{S}ecret \texttt{S}haring based \texttt{M}atrix \texttt{M}ultiplication (\texttt{SSMM}) protocol to optimize \texttt{SeSoRec}, and we also prove its correctness and security. 
\item Our proposed \texttt{SeSoRec} and \texttt{SSMM} have linear computation and communication complexities. 
\item Experimental results conducted on three real-world datasets demonstrate the effectiveness of  \texttt{SeSoRec} and \texttt{SSMM}. 
\end{itemize}
\section{Background}
In this section, we review related backgrounds, including (1) social recommendation, (2) secure multi-party computation, and (3) privacy preserving recommendation.

\subsection{Social Recommendation}
Factorization based recommendation \cite{mnih2007probabilistic,koren2009matrix,chen2018distributed,chen2018semi} is one of the most popular approaches in recommender system. 
It factorizes a user-item rating (or other interaction) matrix into a user latent matrix and an item latent matrix. 
However, traditional factorization based approaches assume that users are independent and identically distributed, which is inconsistent with the reality that users are inherently connected via various types of social relations such as friendships and trust relations. 
Therefore, social factorization models incorporate social relationship into account to improve recommendation performance \cite{ma2011recommender}, and the basic intuition is that connected users are likely to have similar preferences. 
According to \cite{tang2013social}, social factorization models can be formally stated as:\\
 \text{social factorization model} = \text{basic factorization model} + social information model. 

To date, different social information models were proposed to capture social information, and the basic intuition is that connected users are likely to have similar preferences. 

\subsection{Secure Multi-Party Computation}
The concept of secure Multi-Party Computation (MPC) was formally introduced in \cite{yao1982protocols}, which aims to generate methods (or protocols) for multi-parties to jointly compute a function (e.g., vector multiplication) over their inputs (e.g., vectors for each party) while keeping those inputs private. 
MPC can be implemented using different protocols, such as garbled circuits \cite{yao1986generate}, GMW \cite{goldreich1987play}, and secret sharing \cite{shamir1979share}. 
MPC has been applied into many machine learning algorithms, such as decision tree \cite{lindell2005secure}, linear regression \cite{nikolaenko2013privacy}, logistic regression \cite{mohassel2017secureml}, and collaborative filtering \cite{shmueli2017secure}. 
In this paper, we propose a secret sharing based matrix muliplication algorithm for secure social recommendation. 

\subsection{User Privacy Preserving Recommendation}
Another related research area belongs to privacy preserving recommendation. 
Recently, user privacy has drawn lots of attention, and how to train models while keeping user privacy becomes a hot research topic, e.g., \textit{federated learning} and \textit{shared machine learning} \cite{mcmahan2016communication,chen2018}. 
There are research works adopt garbled circuits to protect user privacy while making recommendation \cite{nikolaenko2013privacymf}. 
Some other works use differential privacy to protect user privacy while training recommendation models \cite{mcsherry2009differentially,hua2015differentially,meng2018personalized}.

\nosection{Difference between user privacy and data security}
User privacy preserving recommendation aims to protect user privacy on the customer side (2C), while data security based recommendation intends to protect the data security of business partners who have already collected users' private data (2B). 
In this paper, we aim to (1) integrate rating platform and social platform for better recommendation, 
and (2) protect the data security of both platforms.

\section{The Proposed Model}\label{model}
In this section, we first formally describe the secure social recommendation problem, and then present our proposed \texttt{SE}cure \texttt{SO}cial \texttt{REC}ommendation (\texttt{SeSoRec})  framework for this problem. 

\subsection{Problem Definition}
Formally, let $\mathcal{A}$ be the user-item interaction platform, and $\mathcal{U}$ and $\mathcal{V}$ be the user and item set on it, with $I$ and $J$ denoting user size and item size, respectively. 
Let $\mathcal{R}$ be user-item interaction set between user $i \in \mathcal{U}$ and item $j \in \mathcal{V}$, $|\mathcal{R}|$ is the total number of ratings.
Let $\textbf{R}$ be the user-item interaction matrix, with element $r_{ij}$ being the rating of user $i$ on item $j$. 
Let $\textbf{U}\in \mathbb{R}^{K\times{}I}$ and $\textbf{V}\in \mathbb{R}^{K\times{}J}$ denote the user and item latent factor matrices, with their column vectors $\textbf{u}_\textbf{i}$ and $\textbf{v}_\textbf{j}$ being the $K$-dimensional latent factors for user $i$ and item $j$, respectively. 
Let $\mathcal{B}$ be the user social platform, and we assume that the social platform $\mathcal{B}$ has the same user set $\mathcal{U}$ as the user-item interaction platform $\mathcal{A}$. 
We further let $\textbf{S}$ be the user-user social matrix\footnote{Note that our model can be slightly modifed to meet the case when $\textbf{S}$ is asymmetric.}
, with the element $s_{if}$ being the social relationship strength between user $i$ and user $f$. 


The problem of secure social recommendation is that, platforms $\mathcal{A}$ and $\mathcal{B}$ securely keep their own data and model, meanwhile $\mathcal{A}$ can improve its recommendation performance by utilizing the social information of $\mathcal{B}$.

\subsection{Secure Social Recommendation Framework}
Social recommendation can be formalized as a basic factorization model plus a social information model, based on the assumption that connected users tend to have similar preferences, as described in Section 2.1. 
Most existing social factorization models have the following objective function 
\begin{equation}
\label{social}
\begin{split}
\mathop { \min }\limits_{\textbf{u}_\textbf{i},\textbf{v}_\textbf{j}} \sum\limits_{i=1}^{I} \sum\limits_{j=1}^{J} f({r_{ij}, \textbf{u}_\textbf{i}, \textbf{v}_\textbf{j}}) 
 + \gamma \sum\limits_{i=1}^{I} \sum\limits_{f=1}^{I} g(s_{if}, \textbf{u}_\textbf{i}, \textbf{u}_\textbf{f}),
\end{split}	
\end{equation}
where $f({r_{ij}, \textbf{u}_\textbf{i}, \textbf{v}_\textbf{j}})$ is the loss of the \textit{basic factorization model} that restricts the relationship between the true ratings and predicted ratings, $g(s_{if}, \textbf{u}_\textbf{i}, \textbf{u}_\textbf{f})$ is the loss of the \textit{social information model} that restricts the preferences of users who have social relations, and $\gamma$ controls the social restriction strength. 
A classical example is the Social Regularizer recommendation (Soreg) approach \cite{ma2011recommender}, where 
\begin{equation}
f({r_{ij}, \textbf{u}_\textbf{i}, \textbf{v}_\textbf{j}})=\frac{1}{2}I_{ij}\left(r_{ij} - {\textbf{u}_\textbf{i}}^T \textbf{v}_\textbf{j}\right)^2,
\end{equation} 
\begin{equation}\label{social-part}
g(s_{if}, \textbf{u}_\textbf{i}, \textbf{u}_\textbf{f})=\frac{1}{2}s_{if} ||\textbf{u}_\textbf{i} - \textbf{u}_\textbf{f} ||_F^2,~~~~~~~
\end{equation} 
where $I_{ij}$ is the indicator function that equals to 1 if there is an existing user-item interaction pair and 0 otherwise, and $||\cdot||_F^2$ is the Frobenius norm. 

Traditional social recommendation frameworks such as Soreg can be efficiently solved by stochastic Gradient Descent (GD). 
However, the social information model in Equation (\ref{social-part}) involves a real number $s_{if}$ which belongs to the social platform $\mathcal{B}$, and two real-valued vectors $\textbf{u}_\textbf{i}$ and $\textbf{u}_\textbf{f}$ which are located on the rating platform $\mathcal{A}$,  
secure computation are not guaranteed due to the breach that $\mathcal{A}$ can easily deduce the values $s_{if}$ belonging to $\mathcal{B}$. 

To solve this problem, we propose to use minibatch GD instead of stochastic GD. 
We use $\textbf{B}$ to denote the user-item rating set in the current minibatch and $|\textbf{B}|$ is the batch size. 
Let $\mathcal{U}_\textbf{B}$ and $\mathcal{V}_\textbf{B}$ be the user set and item set in the current batch, $|\mathcal{U}_\textbf{B}|$ and $|\mathcal{V}_\textbf{B}|$ be the user and item size. Apparently $|\mathcal{U}_\textbf{B}| \le |\textbf{B}|$ and $|\mathcal{V}_\textbf{B}| \le |\textbf{B}|$. 
We use $\textbf{R}_\textbf{B} \in \mathbb{R}^{|\mathcal{U}_\textbf{B}|\times |\mathcal{V}_\textbf{B}|}$ to denote the rating matrix in the current batch, $\textbf{I}_\textbf{B} \in \mathbb{R}^{|\mathcal{U}_\textbf{B}|\times |\mathcal{V}_\textbf{B}|}$ to denote the indicator matrix in the current batch. 
Let $\textbf{U}_\textbf{B} \in \mathbb{R}^{K\times |\mathcal{U}_\textbf{B}|}$ and $\textbf{V}_\textbf{B} \in \mathbb{R}^{K\times |\mathcal{V}_\textbf{B}|}$ be the latent factors of the corresponding users and items in the current minibatch. 
Equation (\ref{social}) becomes 
\begin{equation}
\label{sorec-minibatch}
\begin{split}
\mathop { \min } \limits_{\textbf{U}_\textbf{B},\textbf{V}_\textbf{B}} \mathcal{L}
 &= \frac{1}{2} ||\textbf{I}_\textbf{B} \circ \left( \textbf{R}_\textbf{B} - \textbf{U}_\textbf{B}^T \textbf{V}_\textbf{B} \right)||_F^2 \\
 &+ \frac{\gamma }{2} \textbf{SUM}\left( \textbf{D}_\textbf{B}  \circ (\textbf{U}_\textbf{B}^T \textbf{U}_\textbf{B}) \right) 
 - \gamma \textbf{SUM}\left( \textbf{S}_\textbf{B}  \circ (\textbf{U}_\textbf{B}^T \textbf{U}) \right) \\
 &+ \frac{\gamma }{2} \textbf{SUM}\left( \textbf{E} \circ (\textbf{U}^T \textbf{U}) \right) 
 + \frac{\lambda}{2}\left( {||\textbf{U}_\textbf{B}||_F^2}  +  {||\textbf{V}_\textbf{B}||_F^2}\right),
\end{split}	
\end{equation}
where $\textbf{D}_\textbf{B} \in \mathbb{R}^{|\mathcal{U}_\textbf{B}| \times |\mathcal{U}_\textbf{B}|}$ is a diagonal matrix with diagonal element $d_b=\sum_{f=1}^{I}s_{bf}$, $\textbf{S}_\textbf{B} \in \mathbb{R}^{|\mathcal{U}_\textbf{B}| \times I}$ is the social matrix of the users in current minibatch, and $\textbf{E} \in \mathbb{R}^{I \times I}$ is also a diagonal matrix with diagonal element $e_i=\sum_{b=1}^{|\mathcal{U}_\textbf{B}|}s_{bi}$. 
The gradients of $\mathcal{L}$ in Equation \eqref{sorec-minibatch} with respect to $\textbf{U}_\textbf{B}$ and $\textbf{V}_\textbf{B}$ are 
\begin{equation}
\label{usergradientmini}
\begin{split}
\frac{\partial \mathcal{L}}{\partial \textbf{U}_\textbf{B}} 
&=  -\textbf{V}_\textbf{B} \left( {\left(\textbf{R}_\textbf{B} - \textbf{U}_\textbf{B}^T \textbf{V}_\textbf{B}\right)}^T \circ \textbf{I}_\textbf{B} \right) 
+ \frac{\gamma }{2} \textbf{U}_\textbf{B}\textbf{D}_\textbf{B}^T \\
& ~~~~~-\gamma \textbf{U}\textbf{S}_\textbf{B}^T + \frac{\gamma }{2} \textbf{U}_\textbf{B}\textbf{E}_\textbf{B}^T + \lambda \textbf{U}_\textbf{B}
\end{split}	
\end{equation} 
\begin{equation}
\label{itemgradientmini}
\begin{split}
\frac{\partial \mathcal{L}}{\partial \textbf{V}_\textbf{B}} 
&= -\textbf{U}_\textbf{B} \left( {\left(\textbf{R}_\textbf{B} - \textbf{U}_\textbf{B}^T \textbf{V}_\textbf{B}\right)}^T \circ \textbf{I}_\textbf{B} \right) + \lambda \textbf{V}_\textbf{B}, ~~~~~~
\end{split}	
\end{equation} 
where $\textbf{E}_\textbf{B} \in \mathbb{R}^{|\mathcal{U}_\textbf{B}| \times |\mathcal{U}_\textbf{B}|}$ is a diagonal matrix with diagonal element $e_b=e_{i|i=b}$ which is get by extracting the corresponding users' diagonal elements from $\textbf{E}$ in current batch. 


We observe in Equations \eqref{usergradientmini} and \eqref{itemgradientmini} that the matrix product terms $\textbf{U}_\textbf{B}\textbf{D}_\textbf{B}^T$, $\textbf{U}\textbf{S}_\textbf{B}^T$, and $\textbf{U}_\textbf{B}\textbf{E}_\textbf{B}^T$ are crucial. 
These terms involve one matrix ($\textbf{U}$ or $\textbf{U}_\textbf{B}$) on the rating platform and another matrix ($\textbf{D}_\textbf{B}$, $\textbf{S}_\textbf{B}$, or $\textbf{E}_\textbf{B}$) on the social platform. 
All the other terms can be calculated locally by the rating platform. Therefore, we conclude that \textit{the key to secure social recommendation is the secure matrix multiplication operation,} which is a secure MPC problem. 
We summarize the proposed \texttt{SE}cure \texttt{SO}cial \texttt{REC}ommendation (\texttt{SeSoRec}) solution in Algorithm 1, and will present how to perform secure matrix multiplication in the next section. 



\begin{algorithm}[t]\label{algo-ssrec}
\caption{Secure social recommendation}
\KwIn {The observed rating matrix ($\textbf{R}$) on platform $\mathcal{A}$, user social matrix ($\textbf{S}$) on platform $\mathcal{B}$, regularization strength($\gamma$, $\lambda$), learning rate ($\theta$), and maximum iterations ($T$)}
\KwOut{user latent matrix ($\textbf{U}$) and item latent matrix ($\textbf{V}$) on platform $\mathcal{A}$}

Platform $\mathcal{A}$ initializes $\textbf{U}$ and $\textbf{V}$\\
\For{$t=1$ to $T$}
{
	$\mathcal{A}$ and $\mathcal{B}$ calculate $\textbf{D}^T\textbf{U}$ and $\textbf{S}^T\textbf{U}$ based on the secure matrix multiplication in Algorithm 2\\
	$\mathcal{A}$ locally calculates $\frac{\partial \mathcal{L}}{\partial \textbf{U}}$ based on Equation (\ref{usergradientmini}) \\
	$\mathcal{A}$ locally calculates $\frac{\partial \mathcal{L}}{\partial \textbf{V}}$ based on Equation (\ref{itemgradientmini}) \\
	$\mathcal{A}$ locally updates $\textbf{U}$ by $\textbf{U} \leftarrow \textbf{U} - \theta\frac{\partial \mathcal{L}}{\partial \textbf{U}}$ \\
	$\mathcal{A}$ locally updates $\textbf{V}$ by $\textbf{V} \leftarrow \textbf{V} - \theta\frac{\partial \mathcal{L}}{\partial \textbf{V}}$ \\
}
\Return $\textbf{U}$ and $\textbf{V}$ on $\mathcal{A}$
\end{algorithm}
 
\section{Secret Sharing based Matrix Multiplication}
In this section, we first describe technical preliminaries, and then present a secure  matrix multiplication protocol, followed by its correctness and security proof. 

\subsection{Preliminaries}

\nosection{Secret Sharing}
Our proposal relies on Additive Sharing. 
We briefly review this but refer the reader to \cite{demmler2015aby} for more details.
To additively share $(\textbf{Shr}(\cdot))$ an $\ell$-bit value $x$ for two parties ($\mathcal{A}$ and $\mathcal{B}$), party $\mathcal{A}$ generates $x_\mathcal{B} \in \mathds{Z}_{2^\ell}$ uniformly at random, sends $x_\mathcal{B}$ to to party $\mathcal{B}$, and keeps $x_\mathcal{A}=(x - x_\mathcal{B})$ mod $2^\ell$. We use $\langle x \rangle _i$ to denote the share of party $i$. To reconstruct $(\textbf{Rec}(\cdot, \cdot))$ an additively shared value $\langle x \rangle$, each party $i$ sends $\langle x \rangle _i$ to one who computes $\sum_i x_i$ mod $2^\ell, i \in \{\mathcal{A}, \mathcal{B}\}$. 
In this paper, we denote additive sharing by $\langle \cdot \rangle$. 

\noindent
\textit{Apply to decimal numbers. }
The above protocol can not work directly with decimal numbers, since it is not possible to sample uniformly in $\mathds{R}$ \cite{cock2015fast}. 
Following the existing work \cite{mohassel2017secureml}, we approximate decimal arithmetics by using fixed-point arithmetics. 
First, fixed-point addition is trivial. 
Second, for fixed-point multiplication, we use the following strategy. 
Suppose $a$ and $b$ are two decimal numbers with at most $l_F$ bits in the fractional part,
we first transform them to integers by letting $a'=2^{l_F}a$ and $b'=2^{l_F}b$, and then calculate $z=a'b'$. Finally, the last $l_F$ bits of $z$ are truncated so that it has at most $l_F$ bits representing the fractional part. 
The correctness of the above truncation technique for secret sharing can be found in \cite{mohassel2017secureml}. 

\nosection{Simulation-based Security Proof}
To formally prove that a protocol is secure, we adopt the \textit{semi-honest} point of view \cite{goldreich2004FCV}, where each participant truthfully obeys the protocol while being curious about the other parties' original data. 
Under the \textit{real world} and \textit{ideal world} simulation-based proof \cite{lindell2017simulate}, whatever can be computed by one party can be simulated given only the messages it receives during the protocol, which implies that each party learns nothing from the protocol execution beyond what they can derive from messages received in the protocol. 
To formalize our security proof, we need the following notations:
\begin{itemize}[leftmargin=*] \setlength{\itemsep}{-\itemsep}
\item We use $f(x_1, x_2)$ to denote a function with two variables, where $x_1, x_2\in\{0, 1\}^n$ could be encodings of any mathematical objects, e.g. integers, vectors, matrices, or even functionals. We also use $\pi$ to denote a two-party protocol for computing $f$.

\item The \textit{view} of the $i$-th party ($i\in\{\mathcal{A}$, $\mathcal{B}\}$) during the execution of $\pi$ is denoted as $\textbf{view}_i^{\pi}(x_1, x_2, n)$ which can be expanded as $(x_i, \textbf{r}^i, \textbf{m}^i)$, where $x_i$ is the input of $i$-th party, $\textbf{r}^i$ is its internal random bits, and $\textbf{m}^i$ is the messages received or \textit{derived} by the $i$-th party during the execution of $\pi$. Note that $\textbf{m}^i$ includes all the intermediate messages received, all information derived from the intermediate messages, and also the output of $i$-th party during the protocol. 

\item A \textit{probability ensemble} $X = \{X(a, n)\}_{a\in\{0, 1\}^*; n\in\mathbb{N}}$ is an infinite sequence of random variables indexed by $a\in\{0, 1\}^*$ and $n\in\mathbb{N}$. In the context of secure multiparty computation, $a$ represents each party's input and $n$ represents problem size. 
\end{itemize}

\begin{Definition} 
Two probability ensembles $P = \{P(a, n)\}_{a\in\{0, 1\}^*; n\in\mathbb{N}}$ and $Q = \{Q(a, n)\}_{a\in\{0, 1\}^*; n\in\mathbb{N}}$ are said to be \textit{computatitionally indistinguishable}, denoted by $P\overset{c}{\equiv} Q$, if for every non-uniform polynomial-time algorithm $D$ and every polynomial $p(\cdot)$, there exists an $N\in\mathbb{N}$ such that for every $a\in \{0, 1\}^*$ and every $n\in \mathbb{N}$,
\begin{eqnarray}\nonumber
\left|\Pr\{D(P(a,n)) = 1\}-\Pr\{D(Q(a,n)) = 1\}\right|\leq\frac{1}{p(n)}.
\end{eqnarray}
\end{Definition}

\begin{Definition} \label{def-sim}
Let $f(x_1, x_2)$ be a function. We say a two-party protocol $\pi$ computes $f$ \textit{with information leakage $v_1$ to party $\mathcal{A}$ and $v_2$ to party $\mathcal{B}$} where each party is viewed as semi-honest adversaries, if there exist probabilistic polynomial-time algorithms $\mathcal{S}_1$ and $\mathcal{S}_2$ such that
\begin{eqnarray}\nonumber\small
\{(\mathcal{S}_1(1^n, x_1, v_1(x_1, x_2)))\}_{x_1, x_2, n}\overset{c}{\equiv}\{(\textbf{view}_1^{\pi}(x_1, x_2, n))\},\\
\{(\mathcal{S}_2(1^n, x_1, v_2(x_1, x_2)))\}_{x_1, x_2, n}\overset{c}{\equiv}\{(\textbf{view}_2^{\pi}(x_1, x_2, n))\}, \nonumber
\end{eqnarray}
where $x_1, x_2\in\{0, 1\}^*$ and $|x_1| = |x_2| = n$.
\end{Definition}

\subsection{Secret Sharing based Matrix Multiplication}
Secure matrix multiplication is the key to \texttt{SeSoRec}. 
There are several approaches for secure matrix multiplication, such as homomorphic encryption \cite{han2008privacy,teo2012study,dumas2016private} and the secret sharing scheme \cite{de2017efficient}, among which secret sharing is much more efficient. 
Existing secret sharing based matrix multiplication \cite{de2017efficient} either needs a trusted Initializer (a trusted third party) or expensive cryptographic primitives \cite{keller2018overdrive} to generate randomness before computation, i.e., Beaver’s pre-computed multiplication triplet \cite{beaver1991efficient}. 
We call it Trusted Initializer based Secure Matrix Multiplication (TISMM), which may not be applicable in reality. 
Besides, TISMM needs to generate many random matrices
, causing efficiency concerns. 

In this paper, we propose a novel protocol for secure and efficient matrix multiplication using secret sharing. 
Suppose two parties $\mathcal{A}$ and  $\mathcal{B}$ hold matrix $\textbf{P}\in \mathbb{R}^{x\times{}y}$ and matrix $\textbf{Q}\in \mathbb{R}^{y\times{}z}$ separately, where $y$ is an even number\footnote{One can simply change y to an even number by adding an additional zero column in $\textbf{P}$ and zero row in $\textbf{Q}$}. Our algorithm generalizes the inner product algorithm proposed in \cite{zhu2015ESP} to compute the matrix product $\textbf{PQ}$. 
We first summarize our proposed \texttt{S}ecret \texttt{S}haring based \texttt{M}atrix \texttt{M}ultiplication (\texttt{SSMM}) in Algorithm 2, and then prove its correctness and security.

\subsection{Correctness Proof} 

According to Algorithm 2, we have 
\begin{eqnarray}\small
\label{correctness}
\nonumber
\textbf{M} + \textbf{N} =& (\textbf{P}+2\textbf{P}^\prime)\textbf{Q}_1 + (\textbf{P}_2+\textbf{P}^\prime_o)\textbf{Q}_2 \\
\nonumber
 &+ \textbf{P}_1(2\textbf{Q}-\textbf{Q}^\prime)- \textbf{P}_2(\textbf{Q}_2+\textbf{Q}^\prime_e) \\
 \nonumber
 =& \textbf{P}\textbf{Q}_1 + 2\textbf{P}^\prime\textbf{Q}_1 + \textbf{P}_2\textbf{Q}_2 + \textbf{P}^\prime_o\textbf{Q}_2 \\
 \label{expand}
 &+ 2\textbf{P}_1\textbf{Q}- \textbf{P}_1\textbf{Q}^\prime - \textbf{P}_2\textbf{Q}_2 - \textbf{P}_2\textbf{Q}^\prime_e\\
 \nonumber
 =& \textbf{P}\textbf{Q}^\prime - \textbf{P}\textbf{Q} + 2\textbf{P}^\prime\textbf{Q}^\prime - 2\textbf{P}^\prime\textbf{Q}\\
 \nonumber
 &+ \textbf{P}^\prime_o\textbf{Q}^\prime_e - \textbf{P}^\prime_o\textbf{Q}^\prime_o + 2\textbf{P}\textbf{Q} + 2\textbf{P}^\prime\textbf{Q}\\
 \label{substitute}
 &- \textbf{P}\textbf{Q}^\prime - \textbf{P}^\prime\textbf{Q}^\prime  - \textbf{P}^\prime_e\textbf{Q}^\prime_e - \textbf{P}^\prime_o\textbf{Q}^\prime_e\\
 \label{cancel}
 =& \textbf{P}\textbf{Q} + \textbf{P}^\prime\textbf{Q}^\prime - \textbf{P}^\prime_o\textbf{Q}^\prime_o - \textbf{P}^\prime_e\textbf{Q}^\prime_e\\
 \label{result}
 =&\textbf{P}\textbf{Q}.
\end{eqnarray}
Equation \eqref{substitute} is by substituting $\textbf{P}_1$, $\textbf{P}_2$, $\textbf{Q}_1$, and $\textbf{Q}_2$ in Equation \eqref{expand} according to Algorithm 2 (Line 4 and Line 5). 
Equation \eqref{cancel} holds by simplifying Equation \eqref{substitute}. 
The $(i, j)$-th entry of $\textbf{P}^\prime\textbf{Q}^\prime$ is the inner product of the $i$-th row of $\textbf{P}^\prime$ and the $j$th column of $\textbf{Q}^\prime$. 
Finally, by matrix definition, the $(i, j)$-th entry of $\textbf{P}^\prime_o\textbf{Q}^\prime_o$ (resp. $\textbf{P}^\prime_e\textbf{Q}^\prime_e$) is the inner product of the odd (resp. even) terms in the $i$-th row of $\textbf{P}^\prime$ and the odd (resp. even) terms in the $j$-th column of $\textbf{Q}^\prime$, so we have $\textbf{P}^\prime\textbf{Q}^\prime = \textbf{P}^\prime_o\textbf{Q}^\prime_o + \textbf{P}^\prime_e\textbf{Q}^\prime_e$ and  the last three terms in Line 4 are cancelled. Thus, the correctness is proved.

\subsection{Security Proof} 

\begin{Theorem}\label{thm-smm2}
Protocol of \texttt{SSMM} (Algorithm 2) computes matrix multiplication with information leakage $\textbf{Q}_e-\textbf{Q}_o$ to $\mathcal{A}$ and information leakage $\textbf{P}_e + \textbf{P}_o$ to $\mathcal{B}.$
\end{Theorem}

We first give some intuitive discussions on the information disclosure of Algorithm 2.
Let $\textbf{P}_e$ and $\textbf{P}_o$ be sub-matrices of $\textbf{P}$ constructed by its even columns and odd columns. Similarly let $\textbf{Q}_e$ and $\textbf{Q}_o$ be sub-matrices of $\textbf{Q}$ constructed by its even and odd rows.
As indicated in line 4 of Algorithm 2, $\mathcal{B}$ has $\textbf{P}_1$, $\textbf{P}_2$ from $\mathcal{A}$. By extracting the even column sub-matrix $\textbf{P}^1_e$ and odd column sub-matrix $\textbf{P}^1_o$ of $\textbf{P}_1$, $\mathcal{B}$ can calculate $\textbf{P}_3 = \textbf{P}^1_e + \textbf{P}^1_o $. Since $\textbf{P}_1 =  \textbf{P} + \textbf{P}^\prime$, we have $\textbf{P}^1_e = \textbf{P}_e + \textbf{P}^\prime_e, \textbf{P}^1_o = \textbf{P}_o + \textbf{P}^\prime_o$. Thus,  $\mathcal{B}$ can compute $\textbf{P}_e + \textbf{P}_o$ by subtracting $\textbf{P}_2$ from $\textbf{P}_3$.  Similar arguments will show that $\mathcal{A}$ can compute $\textbf{Q}_e - \textbf{Q}_o$ as partial information obtained from $\mathcal{B}$. Although $\mathcal{A}$ and $\mathcal{B}$ both have some level of information disclosed as discussed above, their own private data are still unrevealed.

We then rigorously prove the security level of SSMM, using the preliminary techniques we have given above. 
Note that we first assume all the matrices are finite field ($\mathds{Z}_{2^\ell}$), and then apply fixed point decimal arithmetics. 
Without loss of generality, we first let $\mathcal{A}$ be the adversary and quantify the information leakage to $\mathcal{B}$.
The view of $\mathcal{A}$ in real world contains all information of matrices $\textbf{P}$ and $\textbf{P}^\prime$ (including their even and odd column sub-matrices), together with $\textbf{Q}_1$ and $\textbf{Q}_2$. The key point of the proof is to construct a simulator which can reproduce the same distribution of $\textbf{Q}_1$ and $\textbf{Q}_2$. 
The simulator $\mathcal{S}_A$ for $\mathcal{A}$'s view proceeds like this: 
\begin{itemize}
\item[1.] Assume $\mathcal{S}_A$ has $\textbf{Q}_e - \textbf{Q}_o$ as prior knowledge;
\item[2.] $\mathcal{S}_A$ generate random matrix $\textbf{Q}^\star \in \mathbb{R}^{y\times{}z}$;
\item[3.] $\mathcal{S}_A$ Calculate $\textbf{Q}^\star_2 = (\textbf{Q}^\star_e - \textbf{Q}^\star_o) - (\textbf{Q}_e - \textbf{Q}_o)$.
\end{itemize}

\noindent
$\textbf{Q}^\star_e$ and $\textbf{Q}^\star_o$ are similarly defined as the even column and odd column sub-matrices of $\textbf{Q}^\star$. We claim that $(\textbf{Q}^\star, \textbf{Q}^\star_2)$ has the same distribution as $(\textbf{Q}_1, \textbf{Q}_2)$, thus being computationally indistinguishable. To see this, we first notice that $\textbf{Q}_1$ is  the difference between a random matrix $\textbf{Q}^\prime$ and a fixed matrix $\textbf{Q}$, which is equally distributed as a random matrix, say $\textbf{Q}^\star$. With this in mind, it can be seen similarly that $\textbf{Q}_e^\star - \textbf{Q}_e$ is equally distributed as $\textbf{Q}_e^\prime$ and $\textbf{Q}_o^\star - \textbf{Q}_o$ is equally distributed as $\textbf{Q}_o^\prime$. Therefore, $\textbf{Q}^\star_2$ is equally distributed as $\textbf{Q}_2$. Moreover, $\mathcal{S}_A$ can reproduce all information of matrices $\textbf{P}$ and $\textbf{P}^\prime$. 
So with additional information of $\textbf{Q}_e - \textbf{Q}_o$, the ideal world simulator $\mathcal{S}_A$ successfully reconstructs the view of $\mathcal{A}$, which is equivalent to say that in the real world, only partial information $\textbf{Q}_e - \textbf{Q}_o$ has been disclosed to $\mathcal{A}$ after running the protocol. 

Similar simulator $\mathcal{S}_B$ can be constructed when assuming $\mathcal{B}$ as the adversary. This completes the security proof.

\begin{algorithm}[t]\label{algo-ssmm}
\caption{\texttt{S}ecret \texttt{S}haring based \texttt{M}atrix \texttt{M}ultiplication (\texttt{SSMM})}
\KwIn {A private matrix $\textbf{P}\in \mathbb{R}^{x\times{}y}$ for $\mathcal{A}$, and a private matrix $\textbf{Q}\in \mathbb{R}^{y\times{}z}$ for $\mathcal{B}$}
\KwOut{A matrix $\textbf{M}\in \mathbb{R}^{x\times{}z}$ for $\mathcal{A}$, and a matrix $\textbf{N}\in \mathbb{R}^{x\times{}z}$ for $\mathcal{B}$, such that $\textbf{M} + \textbf{N} = \textbf{P}\textbf{Q}$}

$\mathcal{A}$ and $\mathcal{B}$ locally generate random matrices $\textbf{P}^\prime \in \mathbb{R}^{x\times{}y}$ and $\textbf{Q}^\prime \in \mathbb{R}^{y\times{}z}$ \\
$\mathcal{A}$ locally extracts even columns and odd columns from $\textbf{P}^\prime$, and get $\textbf{P}^\prime_e\in \mathbb{R}^{x\times{}\frac{y}{2}}$ and $\textbf{P}^\prime_o\in \mathbb{R}^{x\times{}\frac{y}{2}}$ \\
$\mathcal{B}$ locally extracts even rows and odd rows from $\textbf{Q}^\prime$, and get $\textbf{Q}^\prime_e\in \mathbb{R}^{\frac{y}{2}\times{}z}$ and $\textbf{Q}^\prime_o\in \mathbb{R}^{\frac{y}{2}\times{}z}$ \\
$\mathcal{A}$ computes $\textbf{P}_1=\textbf{P} + \textbf{P}^\prime$ and $\textbf{P}_2=\textbf{P}^\prime_e + \textbf{P}^\prime_o$, and sends $\textbf{P}_1$ and $\textbf{P}_2$ to $\mathcal{B}$ \\
$\mathcal{B}$ computes $\textbf{Q}_1=\textbf{Q}^\prime - \textbf{Q}$ and $\textbf{Q}_2=\textbf{Q}^\prime_e - \textbf{Q}^\prime_o$, and sends $\textbf{Q}_1$ and $\textbf{Q}_2$ to $\mathcal{A}$ \\
$\mathcal{A}$ locally computes $\textbf{M} = (\textbf{P}+2\textbf{P}^\prime)\textbf{Q}_1 + (\textbf{P}_2+\textbf{P}^\prime_o)\textbf{Q}_2$\\
$\mathcal{B}$ locally computes $\textbf{N} = \textbf{P}_1(2\textbf{Q}-\textbf{Q}^\prime) - \textbf{P}_2(\textbf{Q}_2+\textbf{Q}^\prime_e)$\\
$\mathcal{B}$ sends $\textbf{N}$ to $\mathcal{A}$, and $\mathcal{A}$ calculates $\textbf{M} + \textbf{N}$\\
\Return $\textbf{M} + \textbf{N}$ for $\mathcal{A}$
\end{algorithm}

\noindent
\textbf{Complexity Analysis of \texttt{SSMM}.} The computational complexity mainly comes from Line 6 and 7 in Algorithm 2, which is $O(x\times y\times z)$. The communication complexity from $\mathcal{A}$ to $\mathcal{B}$ depends on matrices $\textbf{P}_1$ and $\textbf{P}_2$, both of which are $O(x\times y)$. 
The communication complexity from $\mathcal{B}$ to $\mathcal{A}$ depends on matrices $\textbf{Q}_1$, $\textbf{Q}_2$, and $\textbf{N}$, which are $O(y\times z)$, $O(y\times z)$, and $O(x\times z)$, respectively, and $O((x+y)\times z)$ in total. 

When one of the matrices is sparse, we can slightly modify the secret sharing strategy in Algorithm 2 such that both the computational and communication complexities are reduced accordingly. 
Without loss of generality, we assume $\textbf{Q}$ is sparse in the sense that for the rows in $\textbf{Q}$ the average number of non-zero entries is $d\ll z$. When generating $\textbf{Q}^\prime$, $\mathcal{B}$ does not make it so dense as in Line 1 of Algorithm 2. 
The new strategy for generating $\textbf{Q}^\prime$ is as follows:

\For{each row in $\textbf{Q}$}
{
	\begin{itemize}[leftmargin=*] \setlength{\itemsep}{-\itemsep}
	\item[1.]generate random numbers for all non-zero entries
	\item[2.]randomly select $d' \ll z$ entries from the zero entries and generate random numbers for these entries 
	\end{itemize}
}

The value $d'$ is the selected number of non-zero entries of all rows in $\textbf{Q}$, and the above new strategy makes $d'$ small in order to guarantee that the secret shares from $\mathcal{B}$ to $\mathcal{A}$ are sparse. 
However, as $d'$ becomes smaller, $\mathcal{A}$ would obtain more information on $\textbf{Q}$. 
An extremal case is $d' = 0$, in which $\mathcal{A}$ can infer the overall sparsity of $Q$. 
Therefore, a reasonable way is to choose $d' = O(d)$. 
Note that, in practice, one should keep its strategies of choosing $d'$ (i.e., the ratio of $d'/d$ for each row) privately in case of information leakage. 
As long as $\textbf{Q}$ is sparse, $\textbf{Q}_1$ and $\textbf{Q}_2$ are both sparse and can be calculated when generating $\textbf{Q}^\prime$. 
The computational complexity for matrix multiplication decreases to $O(x\times y\times d)$
, and the communication complexity from $\mathcal{B}$ to $\mathcal{A}$ decreases to 
$O(x \times z + y\times d)$, and thus they are significantly reduced compared to the general case analysis.

We remark that the above new secret sharing strategy for sparse matrix exactly satisfy our requirement in \texttt{SeSoRec}. Usually the social matrix $\textbf{S}$ is sparse. When the user social platform $\mathcal{B}$ shares its secrets, it can use the above new strategy to generate its secret shares. 
Moreover, the choice of $d'$ can be private to $\mathcal{B}$ only so that the user-item interaction platform $\mathcal{A}$ cannot gain more information based on the shares from $\mathcal{B}$.
 
\section{Analysis}
In this section, we analyze the time complexity of \texttt{SeSoRec} and discuss its usage and information leakage. 

\subsection{Complexity Analysis of \texttt{SeSoRec}}
We first analyze the communication and computation complexities of \texttt{SeSoRec}, as shown in Algorithms 1. 
Recall that $I$ is user number, $|\mathcal{U}_\textbf{B}|$ and $|\mathcal{V}_\textbf{B}|$ denote the user and item numbers in the current minibatch respectively, $K$ denotes the dimension of latent factor, and $|\mathcal{R}|$ is the number of ratings (data zise). 

\noindent
\textbf{Communication Complexity. }
The communications come from the calculations of $\textbf{U}_\textbf{B}\textbf{D}_\textbf{B}^T$, $\textbf{U}\textbf{S}_\textbf{B}^T$, and $\textbf{U}_\textbf{B}\textbf{E}_\textbf{B}^T$ using \texttt{SSMM}. 
First, for $\textbf{U}_\textbf{B}\textbf{D}_\textbf{B}^T$ and $\textbf{U}_\textbf{B}\textbf{E}_\textbf{B}^T$, by refering to the complexity analysis of the modified \texttt{SSMM}, their communication costs are both $O(|\mathcal{U}_\textbf{B}| \times |\mathcal{U}_\textbf{B}|)$ for each minibatch, and are both $O(|\mathcal{R}| / |\textbf{B}| \times |\mathcal{U}_\textbf{B}| \times |\mathcal{U}_\textbf{B}|) \le O(|\mathcal{R}| \times |\textbf{B}|)$ for passing the dataset once. 
Seconed, for $\textbf{U}\textbf{S}_\textbf{B}^T$, the communication of $\textbf{U}$ only needs to be done once for each data pass, and therefore, its communication cost is $O(I \times K)$. 
To this end, the total communication costs are $O(|\mathcal{R}| \times |\textbf{B}|)+O(I \times K)$ for passing dataset once. 
Since, $|\textbf{B}| \ll |\mathcal{R}|$ and $K \ll I < |\mathcal{R}|$, the total communication cost is linear with data size. 

\noindent
\textbf{Computation Complexity. }
Suppose the average number of neighbors for each user on platform $\mathcal{B}$ is $|\mathcal{N}|$. 
The time complexity of lines 6 and 7 in Algorithm 2 is $O(|\mathcal{U}_\textbf{B}| \times |\mathcal{N}| \times K)$ for each minibatch, and is $O(|\mathcal{R}| / |\textbf{B}| \times |\mathcal{U}_\textbf{B}| \times |\mathcal{N}| \times K) \le O(|\mathcal{R}| \times |\mathcal{N}| \times K)$ for passing the dataset once.
Similarly, the time complexity of the lines 3 and 4 in Algorithm 1  for passing the dataset once is $O(|\mathcal{R}| / |\textbf{B}| \times |\mathcal{U}_\textbf{B}| \times |\mathcal{V}_\textbf{B}| \times K) \le O(|\mathcal{R}| \times |\textbf{B}| \times K)$. 
Since $|\mathcal{N}|, |\textbf{B}|, K \ll |\mathcal{R}|$, the total computation cost is also linear with data size. 

By applying minibatch gradient descent, the communication and computation complexities of \texttt{SeSoRec} are both linear with data size and thus can scale to large dataset. 

\subsection{Discussion}\label{discussion}

\textbf{Secure common user identification.} 
Our proposed \texttt{SeSoRec} assumes that platforms $\mathcal{A}$ and $\mathcal{B}$ have the same user set in common, so that they can proceed \texttt{SSMM}. 
The essence of secure common user identification is \textit{private set intersection} (PSI). 
Existing work \cite{pinkas2014faster} has provided efficient solution. 
PSI can be applied to identify common users on two platforms privately before adopting \texttt{SeSoRec} in practice, which guarantees that nothing reveals but the IDs of common users. 

\noindent
\textbf{Information leakage.} 
\texttt{SeSoRec} is asymmetric for two parties, that is, the rating platform $\mathcal{A}$ and the social platform $\mathcal{B}$ collaboratively conduct \texttt{SSMM} and return the results to $\mathcal{A}$. 
Therefore, $\mathcal{B}$ reveals more information to $\mathcal{A}$. 
Although we have proven its security, it may still cause information leakage of $\mathcal{B}$ when $\mathcal{A}$ maliciously initiate \texttt{SSMM} iteratively. 
Suppose $\mathcal{A}$ and $\mathcal{B}$ calculate $\textbf{P}\textbf{Q}$ using \texttt{SSMM}, $\mathcal{A}$ can infer $\textbf{Q}$ by varying $\textbf{P}$ and fixing $\textbf{Q}$ and doing this procedure with enough rounds. 
A naive solution is to set a constraint on $\textbf{Q}$ when conducting \texttt{SSMM}. 
As long as $\textbf{Q}$ (users in each minibatch) is different in each iteration, \texttt{SeSoRec} will have no information leakage. 
We leave better solutions of this as a future work. 
Moreover, when one matrix is sparse in \texttt{SSMM} and the strategies of choosing $d'$ are exposed, the social platform $\mathcal{B}$ may leak some social information to $\mathcal{A}$. 
Specifically, under this circumstance, the sparsity of the social matrix on $\mathcal{B}$ is leaked to $\mathcal{A}$, however the specific social values are still protected. 
Therefore, it is crucial that $\mathcal{B}$ keeps its selection of $d'$ for each row of the social matrix privately.

\section{Experiments}\label{experiments}
In this section, we perform experiments to answer the following question. 
\textbf{Q1:} how does \texttt{SeSoRec} perform comparing with the classic matrix factorization and unsecure social recommendation models, 
\textbf{Q2:} what is the performance of \texttt{SSMM} comparing with the existing TISMM, and
\textbf{Q3:} how does the social parameter ($\lambda$) affect our model performance. 

\subsection{Setting}
We first describe the datasets, metrics, and comparison methods we use in experiments. 

\noindent
\textbf{Datasets.} 
We use three public real-world datasets, i.e., \emph{Epinions} \cite{massa2007trust}, \emph{FilmTrust} \cite{guo2013novel} and \emph{Douban} Movie \cite{Zhong2012CAT}. 
All these datasets contain user-item ratings and user social (trust) information, and are widely adopted in literature. 
Note that although rating and social information are both available in these datasets, we realistically assume that they are located on separate platforms without any possibility of data sharing, which has no side-effect on experiments.


\begin{table}
\centering
\caption{Dataset statistics. Assuming that rating information exist on $\mathcal{A}$ and social information are available on $\mathcal{B}$.}
\begin{tabular}{|c|c|c|c|c|}
  \hline
  Dataset & \#user & \#item & \#rating($\mathcal{A}$) & \#social($\mathcal{B}$) \\
  \hline
  \hline
  \emph{Epinions} & 8,619 & 5,539 & 229,920 & 232,461 \\
  \hline
  \emph{FilmTrust} & 1,508 & 2,071 & 35,497 & 1,853 \\
  \hline
  \emph{Douban} & 13,530 & 13,363 & 2,530,594 & 264,811 \\
  \hline
\end{tabular}
\label{dataset}
\end{table}

Since the original rating matrices of \emph{Epinions} and \emph{Douban} are too sparse, we filter out the users and items whose interactions are less than 20. 
Table \ref{dataset} shows the statistics of these datasets after preprocessing, with which we use \textit{five-fold cross validation} method to conduct experiments and evaluate model performance. 
That is, we split the dataset into five parts, and each time we use four parts as the training set and take the last part as test set. 

\noindent
\textbf{Metrics.} 
To evaluate model performance, we adopt two types of metrics, Root Mean Square Error (RMSE) and Normalized Discounted Cumulative Gain (NDCG@n), both of which are popularly used to evaluate factorization based recommendation performance in literature \cite{koren2009matrix,he2015trirank}. 
RMSE is defined as 
\begin{equation}\small\nonumber
RMSE=\sqrt{\frac{1}{|\tau|}\sum\limits_{(i,j) \in \tau}{(r_{ij}-\hat{r_{ij}})}^2},
\end{equation}
where $\hat{r_{ij}}$ is the predicted rating of user $i$ on item $j$, and $|\tau|$ is the number of predictions in the test dataset $\tau$. 
RMSE evaluates the error between real ratings and predicted ratings, with smaller values indicating better performance. 
NDCG@n is defined as
\begin{equation}\nonumber\small
	NDCG@n=Z_n\sum\limits_{n'=1}^{n}\frac{2^{r_n'}-1}{log_2(n'+1)},
\end{equation} 
where $Z_n$ is a normalizer to ensure that the perfect ranking has value 1 and $r_n'$ is the relevance (real ratings) of item at position $n'$. 
NDCG evaluates the ranking performance of recommendation models, with larger values being better. 
We report NDCG@10 in experiments, and abbreviate it as NDCG.

\noindent
\textbf{Comparison methods.} 
Our proposed \texttt{SeSoRec} is a novel secure social recommendation model, which is a secure version of Soreg \cite{ma2011recommender}. 
We compare \texttt{SeSoRec} with the following latent factor models:

\begin{itemize}[leftmargin=*] \setlength{\itemsep}{-\itemsep}
    \item \textbf{MF} \cite{mnih2007probabilistic} is a classic latent factor model, which only uses the user-item interaction information on platform $\mathcal{A}$. 
    This is the situation where the social platform $\mathcal{B}$ is reluctant to share raw social information with the rating platform $\mathcal{A}$.
    \item \textbf{Soreg} \cite{ma2011recommender} is a classic social recommendation model, which is unsecure in the sense that $\mathcal{A}$ needs the raw data of $\mathcal{B}$. 
    
\end{itemize}

Note that we do not compare with the state-of-the-art recommendation methods. 
The reason is: (1) most of them assume the recommendation platform has many different kinds of information such as contextual information \cite{rendle2010factorization}, which are unfair for our method to compare with, and 
(2) our focus is to study the difference between traditional unsecure social recommendation models and our proposed secure social recommendation model. 


\begin{table}
\centering
\caption{Performance comparison on three dataset, including RMSE and NDCG.}
\begin{tabular}{|c|c|c|c|c|}
 \hline
  Dataset & Metrics & MF & Soreg & \texttt{SeSoRec}\\
  \hline
  \hline
  \multirow{ 2}{*}{Epinions} & RMSE & 1.2687 & 1.1791 & 1.1789 \\
   & NDCG & 0.0363 & 0.0405 & 0.0401 \\
  \hline
  \hline
   \multirow{ 2}{*}{FilmTrust} & RMSE & 1.1907 & 1.1754 & 1.1752 \\  
   & NDCG & 0.2042 & 0.2128 & 0.2124 \\
  \hline
  \hline
  \multirow{ 2}{*}{Douban} & RMSE & 0.7489 & 0.7420 & 0.7419 \\
   & NDCG & 0.0749 & 0.0780 & 0.0778 \\
  \hline
\end{tabular}
\label{compare}
\end{table}

\begin{table}
\centering
\caption{Running time comparison of \texttt{SSMM} and TISMM. }
\begin{tabular}{|c|c|c|c|c|}
  \hline
  dimension ($h$) & 100 & 1000 & \ 10000 \\
  \hline
  \hline
  \texttt{SSMM} & 0.0025 & 0.3246 & 40.744 \\
  \hline
  TISMM & 0.0060 & 0.7279 & 105.83 \\
  \hline
\end{tabular}
\label{exp-ssmm}
\end{table}


\noindent
\textbf{Hyper-parameters.} 
We set the latent factor dimension $K=10$, batch size $|\textbf{B}|=64$, and vary regularizer $\lambda$ and learning rate $\theta$ to choose their best values. 
We also vary $\gamma$ in $\{10^{-2},10^{-1},10^{0},10^{1}\}$ to study its effects on \texttt{SeSoRec}. 
For other parameters, e.g., regularizer $\lambda$ and learning rate $\theta$, we use grid search to find their best values of each model. 

\subsection{Comparison Results (To \textbf{Q1})}
We report the comparison results on three datasets in Table \ref{compare}. From it, we can observe that: 
(1) Soreg and \texttt{SeSoRec} consistently outperform MF. Moreover, we find that the sparser the dataset is, the more Soreg and \texttt{SeSoRec} improve MF. 
  Take RMSE for example, \texttt{SeSoRec} improves MF at 7.60\%, 1.3\%, and 0.98\% on \emph{Epinions}, \emph{FilmTrust}, and \emph{Douban}, with their rating densities 0.48\%, 1.14\%, and 1.4\%, respectively. 
  The results prove that social information is indeed important to recommendation performance, especially when data is sparse. 
(2) Soreg and \texttt{SeSoRec} achieve almost the same recommendation accuracy, where the differences come from the fixed point decimal numbers in secret sharing. The result further validates the correctness of our proposed \texttt{SSMM} besides the theoretical proof. 

\subsection{Comparison between \texttt{SSMM} and TISMM (To \textbf{Q2})}
As we described in \texttt{SSMM} section, existing Trusted Initializer based Secure Matrix Multiplication (TISMM) \cite{de2017efficient} needs a trusted initializer (a trusted third party) to generate secrets before computation. 
Although TISMM may not be applicable in practice, we would like to compare the efficiency of our proposed \texttt{SSMM} with it. 
To this end, we randomly generate two square matrices $\textbf{P} \in \mathbb{R}^{h\times{}h}$ and $\textbf{Q} \in \mathbb{R}^{h\times{}h}$, where $h$ is the dimension of the square matrix. 
We then report the running time (in seconds) of calculating $\textbf{P}\textbf{Q}$ using both algorithms in Table \ref{exp-ssmm}, where we use local area network. 
It can be easily seen that our proposed \texttt{SSMM} costs much less time than TISMM. 
The speedup is around 2.4 times on average. 
This is because TISMM needs to generate more random matrices and involve more matrix operations. 
Moreover, our proposed \texttt{SSMM} protocol does not rely on the trusted initializer which may be difficult to find in practice, thus is more practical. 

\subsection{Parameter Analysis (To \textbf{Q3})}

\begin{figure}[t]
\centering
\subfigure[Effect on RMSE] { \includegraphics[width=3.8cm]{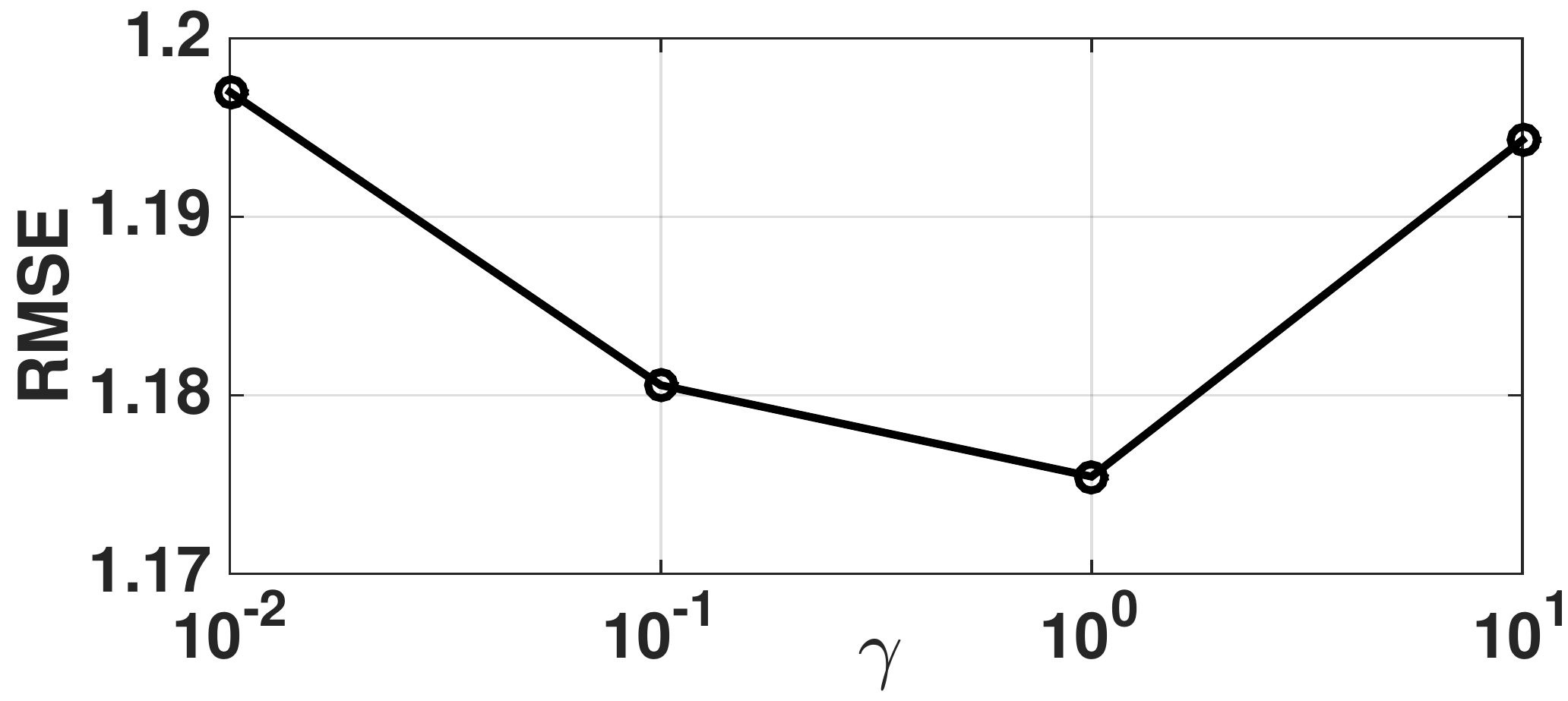}}~~~
\subfigure [Effect on NDCG@10]{ \includegraphics[width=3.8cm]{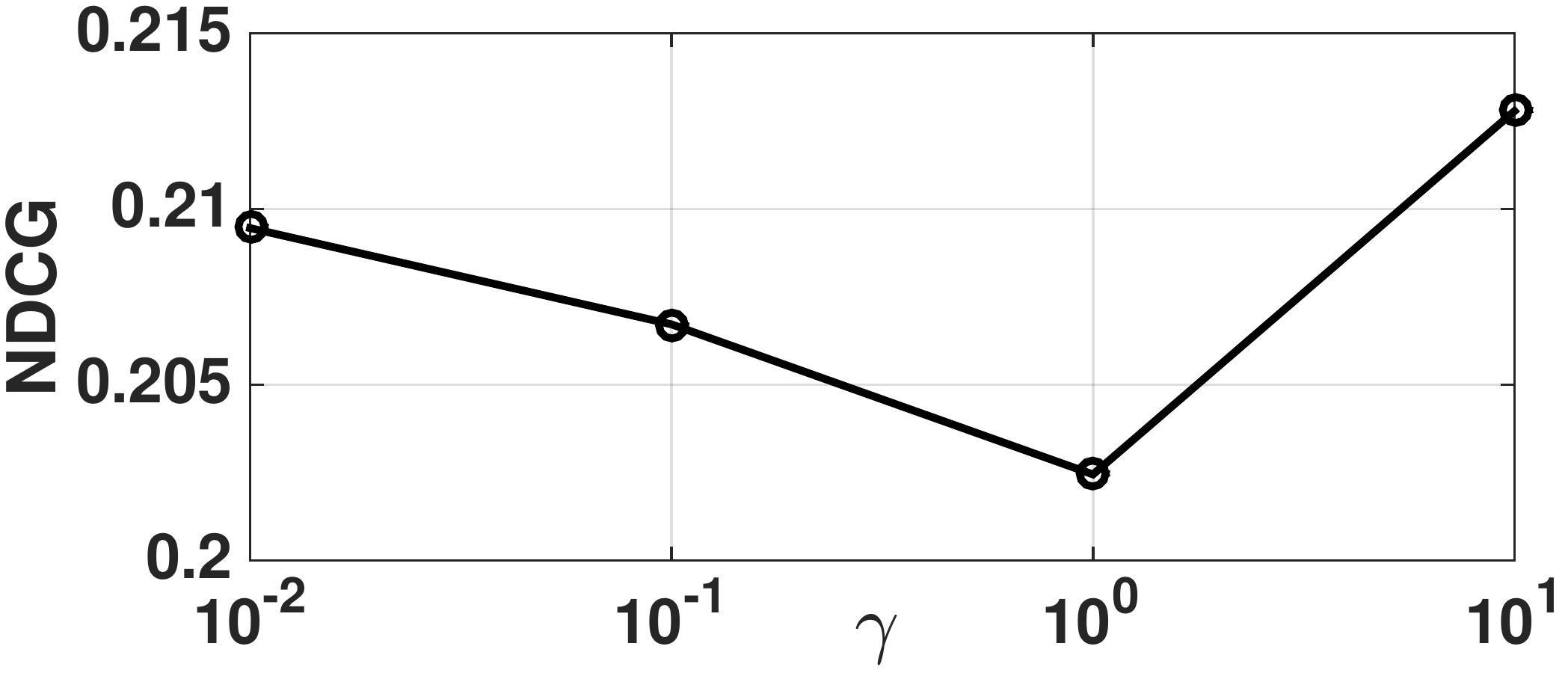}}
\caption{Effect of $\gamma$ on \emph{FilmTrust} dataset.}
\label{effect-ft}
\end{figure}

Finally, we study the effect of social regularizer parameter $\gamma$ on \texttt{SeSoRec}. 
Social recommendation can be formalized as a basic factorization model plus a social information model. 
The social regularizer parameter $\gamma$ controls the contribution of social information model to the final model performance. 
The larger $\gamma$ is, the more likely that the latent factors of connected users are similar, and therefore the more social information model will contribute to the overall performance. 
Figure \ref{effect-ft} shows its effects on \emph{FilmTrust} dataset in terms of both RMSE and NDCG@10. 
It can be seen that with a good choice of $\gamma$, \texttt{SeSoRec} can balance the contribution of user-item rating data on platform $\mathcal{A}$ and user social data on platform $\mathcal{B}$, and thus, achieve the best performance. 


\section{Conclusion and Future Work}

In this paper, we proposed a secret sharing based secure social recommendation framework, which can not only mine knowledge from social platform to improve the recommendation performance of the rating platform, but also keep the raw data of both platforms securely. 
Specifically, we first formalized secure social recommendation as a MPC problem and proposed a \texttt{SE}cure \texttt{SO}cial \texttt{REC}ommendation (\texttt{SeSoRec})  framework for it. 
We then proposed a novel \texttt{S}ecret \texttt{S}haring based \texttt{M}atrix \texttt{M}ultiplication (\texttt{SSMM}) algorithm to optimize it, and proved its correctness and security.
Besides, we analyzed that \texttt{SeSoRec} has linear communication and computation complexities and thus can scale to large datasets. 
Experimental results on real-world datasets demonstrated that, \texttt{SeSoRec} achieves almost the same accuracy as the existing unsecure social recommendation model, and \texttt{SSMM} significantly outperforms the existing trusted initializer based secure matrix multiplication protocol. 
In the future, we would like to solve the potential information leakage problem of \texttt{SeSoRec} with better solutions. 

\bibliographystyle{ecai}
\bibliography{reference}

\end{document}